\documentclass[twoside,11pt]{article}

\usepackage{jmlr2e}
\usepackage{multirow}
\usepackage{amssymb, amsmath}  
\usepackage{color}
\usepackage{url}
\usepackage{algorithm,algcompatible}
\newcommand{\dataset}{{\cal D}}

\newcommand{\M}[1]{\boldsymbol{#1}}  %matrix
\newcommand{\V}[1]{\boldsymbol{#1}}  %vector
\newcommand{\X}[1]{\mathbf{#1}}  %a set of spatial locations

\jmlrheading{XX}{2020}{1-23}{01/20}{XX/XX}{park20a}{Chiwoo Park, David J. Borth, Nicholas S. Wilson, Chad N. Hunter, and Fritz J. Friedersdorf}

\ShortHeadings{Robust Gaussian Process Regression with a Bias Model}{Park, Borth, Wilson, Hunter and Friedersdorf}
\firstpageno{1}

\begin{document}

\title{Robust Gaussian Process Regression with a Bias Model}

\author{\name Chiwoo Park \email cpark5@fsu.edu \\
       \addr Department of Industrial and Manufacturing Engineering\\
       Florida State University \\
       Tallahassee, FL 32310, USA
       \AND
       \name David J. Borth \email David.Borth@udri.udayton.edu \\
       \addr Coatings Corrosion and Erosion Laboratory \\
       University of Dayton Research Institute \\
       Dayton, OH 45469, USA
   \AND
   \name Nicholas S. Wilson \email nicholas.wilson.31@us.af.mil \\
   \addr Materials and Manufacturing Directorate \\
   Air Force Research Laboratory \\
   Wright-Patterson AFB, OH 45433, USA
   \AND
   \name Chad N. Hunter \email chad.hunter@us.af.mil \\
   \addr Materials and Manufacturing Directorate \\
   Air Force Research Laboratory \\
   Wright-Patterson AFB, OH 45433, USA
    \AND
 \name Fritz J. Friedersdorf \email friedersdorf@lunainc.com \\
 \addr  Performance Systems and Analytics \\
 Luna Innovations Incorporated \\
 Charlottesville, VA 22903, USA  
}

\editor{TBD}

\maketitle

\begin{abstract}%   <- trailing '%' for backward compatibility of .sty file
This paper presents a new approach to a robust Gaussian process (GP) regression, creating a non-parametric Bayesian regression estimate robust to outliers. Most existing approaches replace an outlier-prone Gaussian likelihood with a non-Gaussian likelihood induced from a heavy tail distribution, such as the Laplace distribution and Student-t distribution. However, the use of a non-Gaussian likelihood would incur the need for a computationally expensive Bayesian approximate computation in the posterior inferences. The proposed approach models an outlier as a noisy and biased observation of an unknown regression function, and accordingly, the likelihood contains bias terms to explain the degree of deviations from the regression function. We introduce two bias models that handle the bias terms differently, treating a bias as an unknown and fixed quantity or treating a bias as a random quantity. We entail how the biases can be estimated accurately with other hyperparameters by a regularized maximum likelihood estimation. Conditioned on the bias estimates, the robust GP regression can be reduced to a standard GP regression problem with analytical forms of the predictive mean and variance estimates. Therefore, the proposed approach is simple and very computationally attractive. It also gives a very robust and accurate GP estimate for many tested scenarios. For the numerical evaluation, we perform a comprehensive simulation study to evaluate the proposed approach with the comparison to the existing robust GP approaches under various simulated scenarios of different outlier proportions and different noise levels. The approach is applied to data from two measurement systems, where the predictors are based on robust environmental parameter measurements and the response variables utilize more complex chemical sensing methods that contain a certain percentage of outliers.  The utility of the measurement systems and value of the environmental data are improved through the computationally efficient GP regression and bias model. 
\end{abstract}


\section{Introduction}
Regression analysis is a statistical analysis to find an unknown function relating predictor variables to a real response variable given some paired measurements of the predictor and response variables. Typically, the values of the predictor variables are given, and the conditional expectation or probability of the response variable for a given value of the predictor variables is estimated using the given measurements. In practice, the response variable measurements may contain outliers that do not follow the pattern of the other measurements. These outliers can produce a misleading outcome in regression analysis when conventional regression approaches are applied unless the outlier can be taken into account. Robust regression is an approach to overcome this limitation \citep{huber2011robust}. This paper is concerned with a robust regression method.

This work is motivated by the needs of analyzing the time-dependent measurements of environmental factors such as temperature and relative humidity at a specific geographic site over a three month time period using Luna Innovations corrosion and coatings evaluation system ({CorRES}\textsuperscript{\texttrademark}), a multichannel electrochemical monitoring device that continuously records environmental parameters including temperature and relative humidity.  These environmental variables may be used to predict more complex parameters related to atmospheric chemistry that affect degradation of aerospace materials.  A sophisticated atmospheric monitoring system that incorporates gas monitoring sensors manufactured by {Airpointer}\textsuperscript{\textregistered} was used over the same three month time period to obtain response variable measurements for model development.  Data sets from these gas monitoring systems may contain a small percentage of outlier data associated with system startup, user operation, periodic calibration, and other uncontrollable factors such as power fluctuation. Establishing an accurate regression model with less influences from the outliers is crucial for understanding the patterns of the environment factors relevant to degradation of aerospace materials. We are particularly interested in a robust Gaussian process regression with data collected by CorRES and Airpointer monitoring systems, mainly because of the generality and flexibility of a Gaussian process regression.

A Gaussian process (GP) regression is a Bayesian nonparametric approach for regression analysis \citep{rasmussen2006gaussian}. In the approach, a GP prior is placed on an unknown regression function to express the prior belief on the function, and a Gaussian likelihood function is popularly used to model data as the noisy observations of the unknown function with Gaussian noises. It is well known that the Gaussian likelihood is very sensitive to outliers in data in that the mean and variance estimates are significantly affected by the outliers \citep{jylanki2011robust}. This is mainly because the Gaussian density is fast decaying in its tail, so data in the tail part has very low likelihood values.  To address this issue, different likelihood models induced from probability distributions showing heavy-tail behaviors have been tried, including the Student-t likelihood \citep{jylanki2011robust, shah2014student, ranjan2016robust}, Laplace likelihood \citep{kuss2006gaussian, ranjan2016robust},  Gaussian mixture likelihood \citep{naish2008robust, daemi2019identification}, and data-dependent noise model \citep{goldberg1998regression}. Replacing the Gaussian likelihood with a non-Gaussian likelihood makes the derivation of the posterior distribution difficult. Many numerical approximation schemes have been sought with cost of expensive computation steps, e.g., a Markov Chain Monte Carlo sampling and several variants of the Expectation Propagation \citep{minka2001expectation}. 

In this paper, we propose a simpler treatment of outliers. We regard outliers as noisy and biased observations of an underlying regression function, where the `bias' means the deviation of the conditional expectation of the response variable from the regression function. Accordingly, we introduce a bias term in the likelihood that explains the deviation of outliers from a regression trend for a robust GP regression. There is a clear difference from the existing approaches that try to explain outliers with different noise models. In our approach, the unknown bias term is estimated jointly with the hyperparameters of a covariance function of GP. When the bias is estimated, the posterior inference of the unknown regression function follows a standard GP regression solution procedure given in an analytically closed form. This simple idea provides the posterior mean and variance estimates of the regression function in an analytically closed form as the standard GP regression approach, so it is computationally efficient. We will study how the posterior mean and variance estimates behave with the comparisons to the more complex approximation models discussed in the literature. 

The remainder of this paper is organized as follows. Section \ref{section2} describes two bias models to link biased observations to an underlying regression function and entails how the model parameters are estimated jointly with other hyperparameters of the GP regression. We also explain how the robust GP estimates can be achieved with the bias models in the same section. In Section \ref{section3}, we investigate the numerical performance of the proposed method with comparisons to some existing robust GP approaches for various simulation scenarios of different outlier proportions, noise levels, and input dimensions of data. The numerical evaluation is extended with the use of real environmental sensor data containing different patterns of outliers in Section \ref{section4}. We finally conclude this paper in Section \ref{section5}. 

\section{Method} \label{section2}
Consider a robust regression problem of estimating an unknown regression function $f$ that relates a $d$ dimensional predictor $x \in \mathbb{R}^d$ to a real response $y$, using its observations, $\dataset = \{(x_i,y_i), i=1,\ldots, N\}$. The observed response $y_i$ is assumed a noisy and biased observation of $f$,
\begin{equation*}
y_i = \delta_i + f(x_i) + \epsilon_i, \qquad i=1, \dots, N,
\end{equation*}
where $\delta_i$ is the bias of $y_i$ deviating from $f(x_i)$, and $\epsilon_i \sim \mathcal{N}(0, \sigma^2)$ is a random white noise, independent of $f(x_i)$. We regard outliers as data with large bias $\delta_i$. Notice that we do not use bold font for the multivariate predictor $x_i$ and reserve bold font for the collection of observed predictor locations, $\X{x} = [x_1,x_2,\dots,x_N]^T$. Some other notations for the future use are described as follows. Let $\V{y} = [y_1, \ldots, y_N]^T$ denote the corresponding response vector, and let $\V{\delta} = [\delta_1, \delta_2, \ldots, \delta_N]^T$ denote a vector of the unknown bias values.

In GP regression, the unknown function $f$ is assumed a realization of Gaussian process with zero mean and covariance function $c(\cdot, \cdot)$. The covariance function popularly used for $c(\cdot, \cdot)$ is a parametric covariance function such as the Matérn covariance function and the exponential covariance function. We denote the parameters of the covariance by $\V\theta$. Given the GP prior, the observations $\dataset$ is used to obtain the posterior predictive distribution of $f$ at an unobserved location $x_*$, denoted by $f_* = f(x_*)$. The joint distribution of $(f_*, \V{y})$ follows a multivariate normal distribution,
\begin{equation} \label{eq:joint}
P(f_*, \V{y} | \sigma^2, \V{\theta}, \V{\delta}) =
\mathcal{N}\left( \left[ \begin{array} {c} 
0 \\
\V{\delta} \end{array} \right] , \left[ \begin{array} {c c} c_{**}
& \V{c}_{\X{x}*}^T \\ \V{c}_{\X{x}*}
& \sigma^2 \M{I} + \M{C}_{\X{xx}} \end{array} \right] \right),
\end{equation}
where $c_{**} = c(x_*, x_*)$, $\V{c}_{\X{x}*} = (c(x_1,x_*), \dots, c(x_N, x_*))^T$, and $\M{C}_{\X{xx}}$ is an $N \times N$ matrix with $(i,j)^{th}$ entry $c(x_i, x_j)$. The subscripts on $c_{**}, \V{c}_{\X{x}*}$, and $\M{C}_{\X{xx}}$ indicate the two sets of locations between which the covariance is computed, and we have abbreviated the subscript $x_*$ as $*$. 

If the parameters $\sigma^2, \V{\theta}$ and $\V{\delta}$ are given or their estimates $\hat{\V\delta}$, $\hat{\sigma}^2$ and $\hat{\theta}$ are given, we can apply the Gaussian conditioning formula to the joint distribution \eqref{eq:joint} to achieve the posterior predictive distribution of $f_*$ given $\V{y}$,
\begin{equation}\label{eq:pred-dist}
P(f_* | \V{y}, \hat{\sigma}^2, \hat{\theta}, \V{\hat{\delta}})
= \mathcal{N}(
\V{c}_{\X{x}*}^T (\hat{\sigma}^2\M{I} + \M{C}_{\X{xx}})^{-1} (\V{y} - \V{\hat{\delta}}),
c_{**} - \V{c}_{\X{x}*}^T (\hat{\sigma}^2\M{I} + \M{C}_{\X{xx}})^{-1}
\V{c}_{\X{x}*}).
\end{equation}
The predictive mean
$\V{c}_{\X{x}*}^T (\hat{\sigma}^2\M{I} + \M{C}_{\X{xx}})^{-1} (\V{y}- \V{\hat{\delta}})$
is taken to be the point prediction of $f(x)$ at location $x_*$,
and its uncertainty is measured by the predictive variance
$c_{**} - \V{c}_{\X{x}*}^T (\hat{\sigma}^2\M{I} + \M{C}_{\X{xx}})^{-1}
\V{c}_{\X{x}*}$. In the subsections below, we will explain how to estimate the the parameters.  

\subsection{Constant Bias Model}
\begin{figure}
	\includegraphics[width=\textwidth]{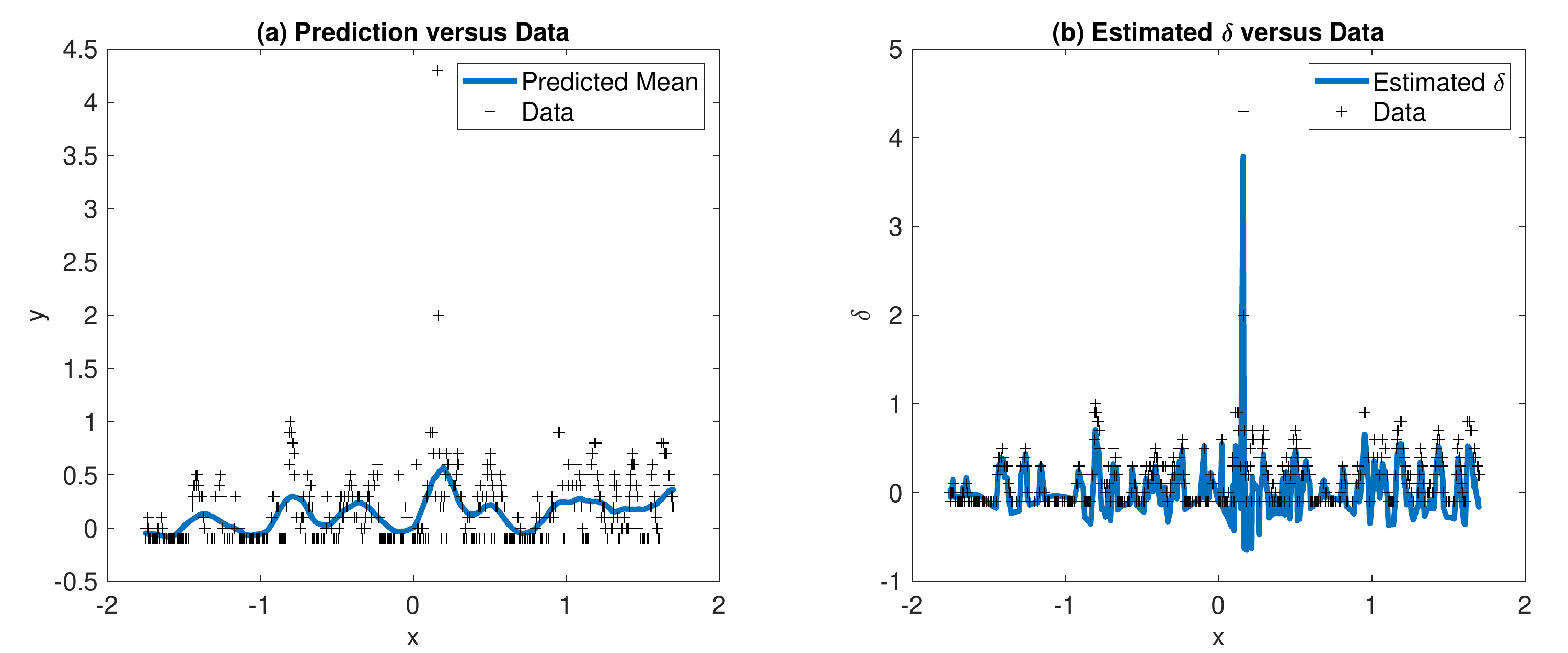}
	\caption{Predicted Mean of $f$ with the ML estimate of $\V{\delta}$. Among 500 data points, two points with $ y > 2$ are outliers. The exponential covariance function was used. The predicted mean is over-smooth mainly because the estimated bias is over-fit to data, absorbing the pattern of $f$ as seen in (b).}
	\label{Figure1}
\end{figure}

We first regard the unknown bias $\V{\delta}$ as a constant vector to estimate, and the likelihood function becomes a Gaussian likelihood with data-dependent means, 
\begin{equation*}
y_i | \delta_i, f, \sigma^2 \sim \mathcal{N}(\delta_i, \sigma^2),
\end{equation*}
which is different from the data-dependent noise variance model \citep{goldberg1998regression}. The bias vector $\V{\delta}$ can be estimated jointly with the other parameters, the noise variance $\sigma^2$ and the covariance parameter $\V{\theta}$. The negative log likelihood function of the parameter is 
\begin{equation*}
\begin{split}
NL_1(\sigma^2, \V{\delta}, \V{\theta})  = & - \log p(\V{y} |\sigma^2, \V{\delta}, \V{\theta}) \\
= & \frac{N}{2}\log (2\pi) + \frac{1}{2} \log |\sigma^2\M{I} + \M{C}_{\X{xx}}| + \frac{1}{2} (\V{y} - \V{\delta})^T (\sigma^2\M{I} + \M{C}_{\X{xx}})^{-1} (\V{y} - \V{\delta}).
\end{split}
\end{equation*}
Estimating the parameters by minimizing the negative log likelihood is not a great idea. First of all, it is not well-defined, since the NL function has infinitely many minima. The estimates $\V{\delta}$ are easily overfit to $\V{y}$, and the $f$ fit to the residual $\V{y}-\V{\delta}$ gives a over-smoothed estimate of $f$ as illustrated in Figure \ref{Figure1}.

\begin{figure}
	\includegraphics[width=\textwidth]{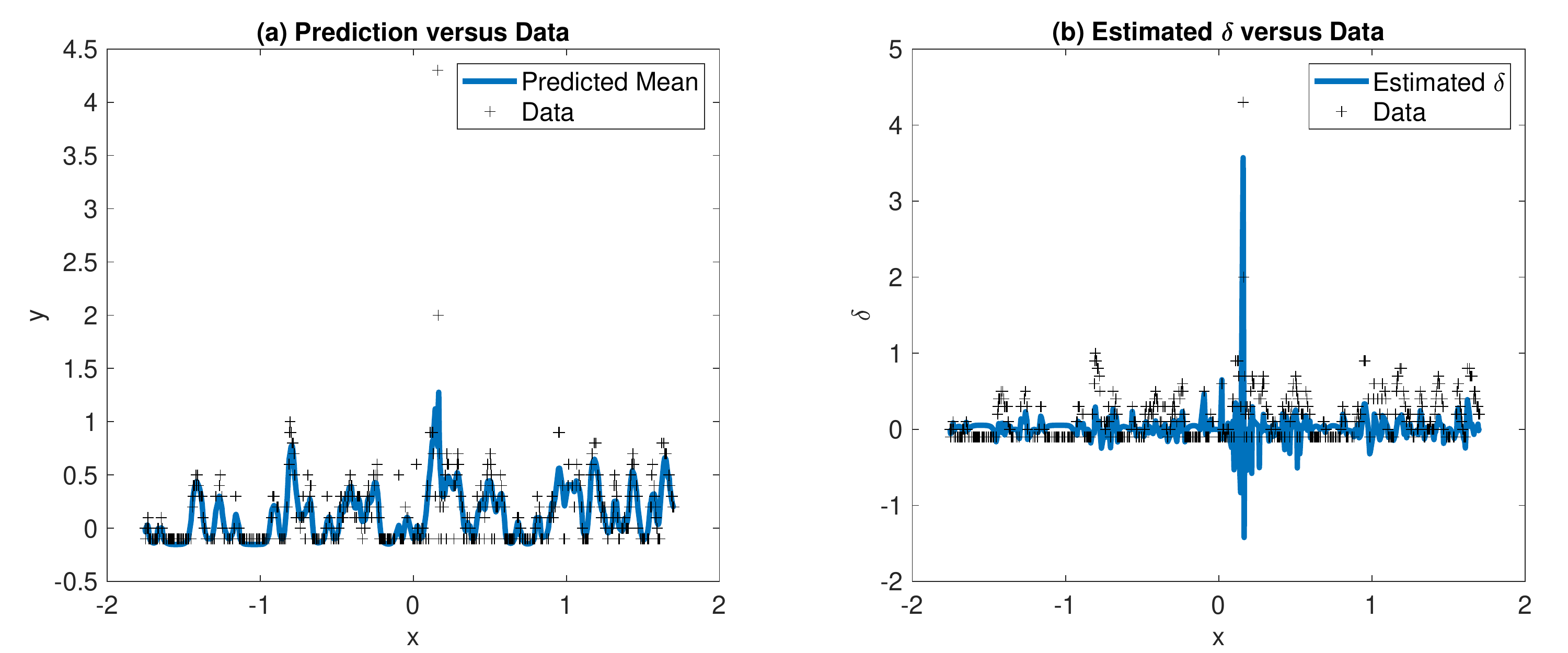}
	\caption{Predicted Mean of $f$ with the Regularized ML estimate of $\V{\delta}$. Among 500 data points, two points with $ y > 2$ are outliers. The exponential covariance function was used. The bias estimates are suppressed near zero except for the outliers.}
	\label{Figure2}
\end{figure}

We place a certain regularization on $\V{\delta}$ for the well-posedness of the problem. We assume that the outliers occur occasionally, so the bias term $\V{\delta}$ should be sparse with many zero elements but very few non-zero elements. We consider to add a $L_1$ regularization on the bias term to the negative log likelihood, 
\begin{equation*}
\begin{split}
RL_1(\sigma^2, \V{\delta}, \V\theta)  = & - \log p(\V{y} |\V\theta) + \lambda |\V{\delta}| \\
= & \frac{N}{2}\log (2\pi) + \frac{1}{2} \log |\sigma^2\M{I} + \M{C}_{\X{xx}}| + \frac{1}{2} (\V{y} - \V{\delta})^T (\sigma^2\M{I} + \M{C}_{\X{xx}})^{-1} (\V{y} - \V{\delta}) + \lambda |\V{\delta}|,
\end{split}
\end{equation*}
where $|\cdot|$ is the $L_1$ norm. The regularized negative log likelihood function is minimized for optimizing $\sigma^2$, $\V{\delta}$ and $\V\theta$ jointly. This regularized ML estimates give a better posterior estimate of $f$ as illustrated in Figure \ref{Figure2}.

In fact, adding the $L_1$ regularization on $\V{\delta}$ to the negative log likelihood function is equivalent to assuming a zero-mean Laplace prior on $\V{\delta}$,
\begin{equation*}
p(\delta_i| \lambda) = \frac{\lambda}{2} \exp\left\{- \lambda|\delta_i| \right\},
\end{equation*}
and then taking a maximum a posteriori probability (MAP) estimate of $\V{\delta}$. Please note that with the Laplace prior, the negative log joint density of $\V{y}$ and $\V{\delta}$ is proportional to $RL_1(\sigma^2, \V{\delta}, \V{\theta})$, 
\begin{equation} \label{eq:JML1}
- \log p(\V{y}, \V{\delta}| \V{\theta}) = RL_1(\sigma^2, \V{\delta}, \V\theta) + \mathcal{R}_1(\lambda),
\end{equation}
where $\mathcal{R}(\lambda) = \log \lambda$ is independent of $\sigma^2, \V{\delta}$ and $\V\theta$. Therefore, minimizing $RL(\sigma^2, \V{\delta}, \theta)$ gives a solution that corresponds to the MAP estimate of $\V{\delta}$. 

\subsection{Random Bias Model}
In this section, we consider the unknown bias $\V{\delta}$ as a random quantity. We first consider a Gaussian prior distribution for the random quantity,
\begin{equation} \label{eq:rbias}
\delta_i \sim \mathcal{N}(\mu_i, \tau_i^2),
\end{equation}
where $\mu_i$ is the mean of the bias, and $\tau_i^2$ is the variance of the bias. In this case, the conditional distribution of $y_i$ given $f$ is
\begin{equation*}
y_i | f, \sigma^2, \mu_i, \tau_i^2 \sim \mathcal{N}(\mu_i, \tilde{\tau}_i^2), 
\end{equation*}
where $\tilde{\tau}_i^2 = \tau_i^2 + \sigma^2$. This over-parameterized model is not a great choice, because the number of the parameters surpasses the number of data. This incurs a non-uniqueness issue in the parameter estimation. 

We consider more restricted models described below. Depending on the restriction, this model would have different likelihoods. For example, 
when $\mu_i = \mu$ and $\tau_i^2 = \tau^2$, the likelihood is reduced to a standard Gaussian likelihood, because
\begin{equation*}
y_i | f, \sigma^2, \mu, \tau^2 \sim \mathcal{N}(\mu, \tilde\tau^2),
\end{equation*} 
where $\tilde\tau^2 = \tau^2 + \sigma^2$. Obviously, we will not consider this case. When $\tau_i^2 = \tau^2$, the likelihood becomes similar to one for the constant bias model which we already discussed in the previous section, because
\begin{equation*}
y_i | f, \sigma^2, \mu_i, \tau^2 \sim \mathcal{N}(\mu_i, \tilde\tau^2). 
\end{equation*} 
When $\mu_i = \mu$, the likelihood is similar to a Gaussian likelihood with data-dependent noise variances, because
\begin{equation*}
y_i | f, \sigma^2, \mu, \tau_i^2 \sim \mathcal{N}(\mu, \tilde{\tau}_i^2). 
\end{equation*}
In this section, we choose this model as a random bias model and discuss the parameter estimation for the model. Please note that, by letting $\mu = 0$, this data-dependent noise variance model is similar to one discussed in \citet{goldberg1998regression}, where $\tilde{\tau}_i^2$ was regressed with a log-Gaussian process prior. Here we treat $\tilde{\tau}_i^2$ as a parameter to estimate. 

\begin{figure}[ht!]
	\includegraphics[width=\textwidth]{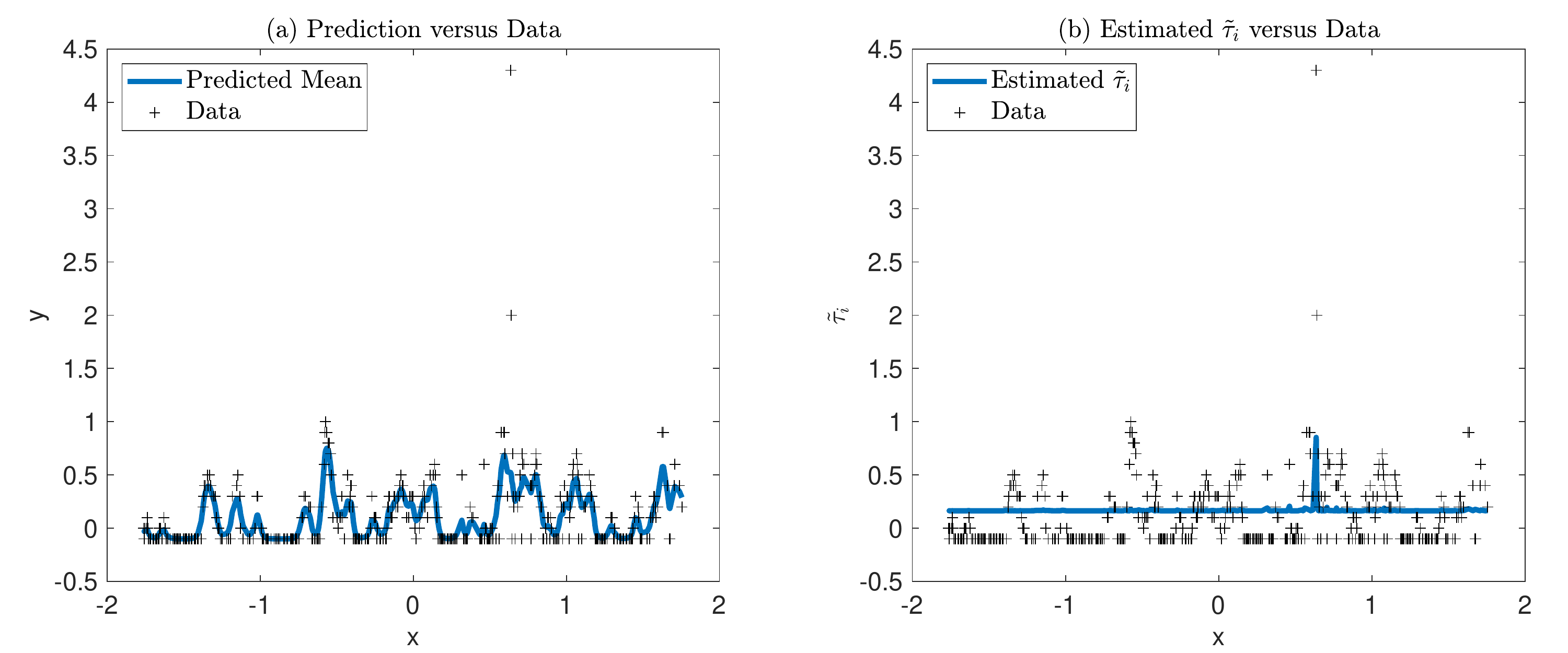}
	\caption{Predicted Mean of $f$ with the Regularized ML estimate of $\V{\tilde{\tau}}$ and $\mu$. Among 500 data points, two points with $y > 2$ are outliers. The exponential covariance function was used. The estimates of $\tilde{\tau}_i$'s are highly concentrated around 0.0268, and they jump to 0.1299 and 0.7276 for the two outliers.}
	\label{Figure3}
\end{figure}

To discuss how we estimate the variance parameters, let $\V{\tilde\tau} = (\tilde{\tau}_1^2, \tilde{\tau}_2^2, \ldots, \tilde{\tau}_N^2)^T$. Given the set of the parameters, the distribution of $\V{y}$ is 
\begin{equation} \label{eq:joint2}
\V{y} | \V{\theta}, \mu, \V{\tilde\tau} \sim
\mathcal{N}\left( \mu\V{1}, \M{D}_{\V{\tilde\tau}} + \M{C}_{\X{xx}} \right),
\end{equation}
where $\M{D}_{\V{\tilde\tau}}$ is a $N \times N$ diagonal matrix with $\tilde{\tau}_i^2$ as its $i$the diagonal element. The negative log likelihood function of the parameter set, $\{\V{\theta}, \mu,  \V{\tilde\tau}\}$, can be
\begin{equation*}
\begin{split}
NL_2(\mu, \V{\theta}, \V{\tilde{\tau}})  = & - \log p(\V{y} |\mu\V{1}, \V{\theta}, \V{\tilde{\tau}}) \\
= & \frac{N}{2}\log (2\pi) + \frac{1}{2} \log |\M{D}_{\V{\tilde\tau}} + \M{C}_{\X{xx}}| + \frac{1}{2} (\V{y} - \mu\V{1})^T (\M{D}_{\V{\tilde\tau}}  + \M{C}_{\X{xx}})^{-1} (\V{y} - \mu\V{1}).
\end{split}
\end{equation*}
Minimizing the negative log likelihood directly for the parameter set can cause the overestimation of the bias parameters, $\mu$ and $\V{\tilde{\tau}}$, leading to the over-smoothed $f$ estimate as we seen in the previous section. We regularize $\mu$ with the $L_2$ norm and regularize each $\log \tilde{\tau}_i^2$ with its linear and negative exponential terms,  
\begin{equation*}
\begin{split}
RL_2(\V{\mu}, \V{\theta}, \V{\tilde{\tau}})  = & NL_2(\mu, \V{\theta}, \V{\tilde{\tau}}) \\
& \quad  + \lambda_1 \mu^2 + \sum_{i=1}^n \lambda_2 \log \tilde{\tau}_i^2 + \lambda_3  \exp(- \log \tilde{\tau}_i^2)  \\
= & \frac{N}{2}\log (2\pi) + \frac{1}{2} \log |\M{D}_{\V{\tilde\tau}} + \M{C}_{\X{xx}}| + \frac{1}{2} (\V{y} - \mu\V{1})^T (\M{D}_{\V{\tilde\tau}}  + \M{C}_{\X{xx}})^{-1} (\V{y} - \mu\V{1}) \\
                                             & \quad + \lambda_1 \mu^2 + \sum_{i=1}^n \lambda_2 \log \tilde{\tau}_i^2 + \lambda_3 \exp(- \log \tilde{\tau}_i^2).
\end{split}
\end{equation*}
The regularization term for $\tilde{\tau}_i^2$ is actually derived from the negative log density of a inverse-gamma distribution. When $\tilde{\tau}_i^2 \sim \mathcal{IG}(\lambda_2-1, \lambda_3)$, its density function is proportional to
\begin{equation*}
g(\tilde{\tau}_i^2)  = \frac{\lambda_3^{\lambda_2-1}}{\Gamma(\lambda_2-1)}\exp\left\{ - \lambda_2 \log \tilde{\tau}_i^2 - \lambda_3 \exp(- \log \tilde{\tau}_i^2)   \right\}.
\end{equation*}
In addition, the regularization term for $\mu$ is proportional to the negative log density of a normal distribution, $\mu \sim \mathcal{N}(0, \lambda_1^{-1}/2)$,
\begin{equation*}
h(\mu) \propto \sqrt{\lambda_1} \exp\{ - \lambda_1 \mu^2 \}. 
\end{equation*} 
The negative log joint density of $\V{y}$, $\mu$ and $\V{\tilde\tau}$ is
\begin{equation} \label{eq:JML2}
-\log p(\V{y}, \mu, \V{\tilde{\tau}}|\V{\theta}) = RL_2(\V{\mu}, \V{\theta}, \V{\tilde{\tau}}) + \mathcal{R}_2(\lambda_1, \lambda_2, \lambda_3),
\end{equation}
where $\mathcal{R}_2(\lambda_1, \lambda_2, \lambda_3) = -\frac{1}{2} \log \lambda_1 - (\lambda_2-1)\log \lambda_3 + \log \Gamma (\lambda_2-1)$. Therefore, minimizing $RL_2$ with respect to the parameter set gives the MAP estimates of $\tilde{\tau}_i$ and $\mu$. The result of the parameter estimation with $RL_2$ is illustrated in Figure \ref{Figure3} with the same toy example which we used in the previous section. The estimates of $\tilde\tau^2_i$'s are highly concentrated on one value except for two outliers, where the variance estimates are significantly higher to mitigate the effect of the outliers on $f$.

\subsection{Tuning Parameter Selection}
Both of the constant bias model and random bias model come with tuning parameters, $\lambda$ and $\{\lambda_1, \lambda_2, \lambda_3\}$. We can choose the tuning parameters iteratively with the regularized ML estimation discussed in the previous section. Let $(\sigma^2_{\lambda}, \V{\delta}_{\lambda}, \V\theta_{\lambda})$ denote the estimates achieved by minimizing $RL_1$ for an initial choice of $\lambda$. We plug in the estimate into the negative log joint density in \eqref{eq:JML1}, 
\begin{equation}
\tilde{\mathcal{R}}_1(\lambda) = RL_1(\sigma^2_{\lambda}, \V{\delta}_{\lambda}, \V\theta_{\lambda}) + \mathcal{R}_1(\lambda).
\end{equation}
We minimize $\tilde{\mathcal{R}}_1(\lambda)$ for updating $\lambda$. This minimization and the minimization of $RL_1$ are iteratively solved until the $\lambda$ value converges. This iteration is actually equivalent to the coordinate descent algorithm to minimize the log joint density, $RL_1(\sigma^2, \V{\delta}, \V\theta) + \mathcal{R}_1(\lambda)$. 

We perform the selection of the tuning parameter $\{\lambda_1, \lambda_2, \lambda_3\}$ similarly. Let $(\mu_{\lambda}, \V{\tilde\tau}_{\lambda}, \V\theta_{\lambda})$ denote the estimates achieved by minimizing $RL_2$ for an initial choice of $\{\lambda_1, \lambda_2, \lambda_3\}$. Plug the estimates into the negative log joint density in \eqref{eq:JML2}, we have
\begin{equation} 
\mathcal{\tilde{R}}_2(\lambda_1, \lambda_2, \lambda_3) = RL_2(\V{\mu}_{\lambda}, \V{\theta}_{\lambda}, \V{\tilde{\tau}}_{\lambda}) + \mathcal{R}_2(\lambda_1, \lambda_2, \lambda_3).
\end{equation}
We minimize $\mathcal{\tilde{R}}_2(\lambda_1, \lambda_2, \lambda_3)$ for updating $\{\lambda_1, \lambda_2, \lambda_3\}$. This minimization and the minimization of $RL_2$ are iteratively solved until the tuning parameter values converge. 

\section{Simulated Examples} \label{section3}
In this section, we present the numerical outcomes of our two proposed bias models with a comprehensive set of simulation scenarios, comparing with the fractional EP approximation with the Student-t likelihood \citep{jylanki2011robust} and the MCMC approach with the Gaussian scale mixture likelihood \citep{gelman2013bayesian}. We use two synthetic datasets, one with one input dimension and another with ten input dimensions. For each dataset, eight simulation scenarios are generated with different outlier patterns, and 15 replicated experiments were performed for each scenario. Below are the details of how we create the scenarios.

The first dataset with one input dimension is generated from the underlying regression function, $$f(x) = 0.3 + 0.4x + 0.5\sin(2.7x) + 1.1/(1+x^2).$$ The 300 training samples, denoted as $\{(x_i, y_i); i = 1,...,300 \}$, are randomly sampled from $x_i \sim \mbox{Unif}(-3, 3)$ and 
\begin{equation*}
y_i = f(x_i) + \epsilon_i.
\end{equation*}
where $\epsilon_i$ is a random variable following a mixture of two Gaussian distributions. We used $\epsilon_i \sim (1-q) \mathcal{N}(0, \sigma^2) + q\mathcal{N}(\mu_o, \sigma^2)$, where $q \in (0, 1)$ quantifies the fraction of outliers, $\mu_o$ is the mean bias of outliers, and $\sigma^2$ is the variance around the mean. We varied the fraction of outliers, $q \in \{0.1, 0.2\}$, and considered $\mu_o \in \{3, 5\}$. We also considered the values $\sigma^2$ to be one sixth of $\mu_o$ or one twelfth of $\mu_o$. If it is the one sixth, the samples from $\mathcal{N}(\mu_o, \sigma^2)$ is slightly overlapped with those from $\mathcal{N}(0, \sigma^2)$, so the outliers are less distinct from the normal data. If it is the one twelfth, they are less likely overlapping. Combining these possible values, we have eight different test scenarios in total. The 1000 test samples, denoted as $\{(x^{(t)}_{i}, y^{(t)}_i)\}$, are randomly sampled from $x^{(t)}_i \sim \mbox{Unif}(-3, 3)$ and $y^{(t)}_i \sim \mathcal{N}(f(x^{(t)}_i), \sigma^2)$.  

The second dataset has ten input dimensions, and the synthetic dataset was first introduced by \citet{friedman1991multivariate} and was referred in many papers for robust GP approaches \citep{kuss2006gaussian}. In the dataset, the underlying regression function $f(x)$ at a 10-dimensional input $x$ depends on the first five input dimensions,  
\begin{equation}
f(x) = 10 \sin (\pi x_1 x_2) + 20 (x_3  - 0.5)^2 + 10x_4 + 5 x_5. 
\end{equation}
The 300 training samples and the 1000 test samples are generated following the same procedure used in the first dataset.  

For each test scenario, we evaluated the four methods, our constant bias model (\texttt{COB}), our random bias model (\texttt{RAB}), the fractional EP approximation with the Student-t likelihood (\texttt{EP}) and the MCMC approach with the Gaussian scale mixture likelihood (\texttt{MCMC}). We collected the total computation time of the four methods to judge their computational efficiency. For comparison of prediction accuracy, we use two measures on the test data, denoted as $\{(x_t, y_t); t=1,\dots,T\}$. The first measure is the mean squared error (MSE)
\begin{equation}
\textrm{MSE} = \frac{1}{T} \sum_{t=1}^T (y_t - \mu_t)^2,
\end{equation}
which measures the accuracy of the mean prediction $\mu_t$ at location $x_t$.
The second one is the negative log predictive density (NLPD)
\begin{equation}
\textrm{NLPD} = \frac{1}{T} \sum_{t=1}^T\left[ \frac{(y_t - \mu_t)^2}{2\sigma_t^2} + \frac{1}{2} \log (2\pi \sigma_t^2) \right],
\end{equation}
which considers the accuracy of the predictive variance $\sigma_t$ as well
as the mean prediction $\mu_t$. These two criteria were used broadly in the GP regression literature. A smaller value of MSE or NLPD indicates a better performance. 

For this numerical analysis, we used the \texttt{GPStuff} Version 4.7 MATLAB package \citep{vanhatalo2013gpstuff} for \texttt{MCMC} and \texttt{EP}. We chose all hyperparameters to be optimized instead of fixing them to the same values. We implemented our two proposed approaches with MATLAB. For all of the four methods, we applied the same squared exponential covariance function. The results with the two datasets are presented in the two subsections below. 

\subsection{1D Synthetic Dataset}
\begin{figure}[ht!]
	\centering
	\includegraphics[width=\textwidth]{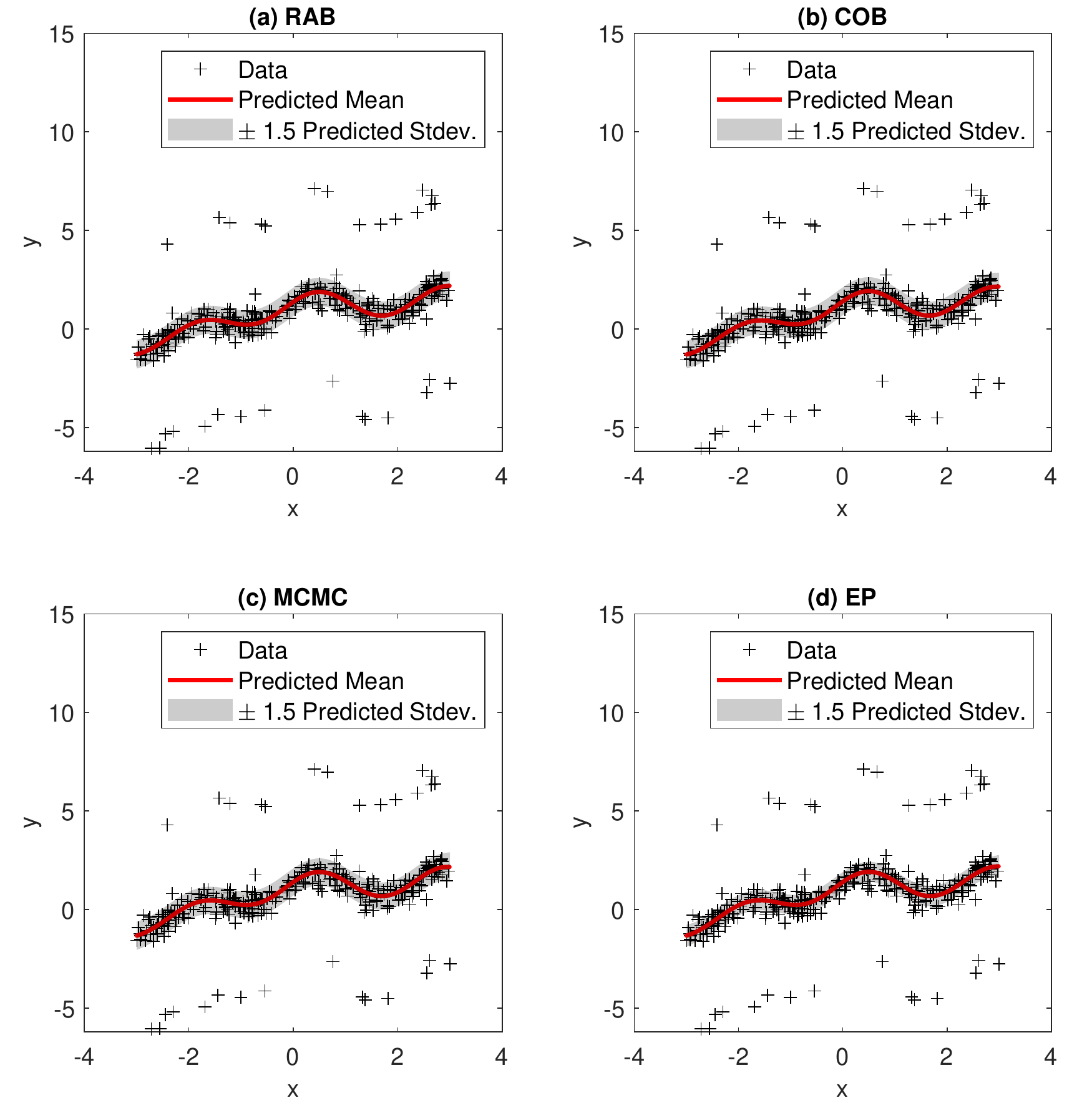}
	\caption{Illustrative outcomes of the four methods compared for $q = 0.1$, $\mu_d = 5$ and $\sigma = \mu_d / 12$. }
	\label{Figure7}
\end{figure}

This 1D example is mainly used to illustrate the outcomes of the four compared methods and compare the predictive means and variances of the four compared methods in a low dimensional case. Figure \ref{Figure7} shows the outcomes of the four methods compared for $q = 0.1$, $\mu_d = 5$ and $\sigma = \mu_d / 12$. All of the four methods work comparably while the \texttt{EP} approach underestimates the predictive variance for this example, giving narrower confidence intervals. 

\begin{figure}[p]
	\includegraphics[width=\textwidth]{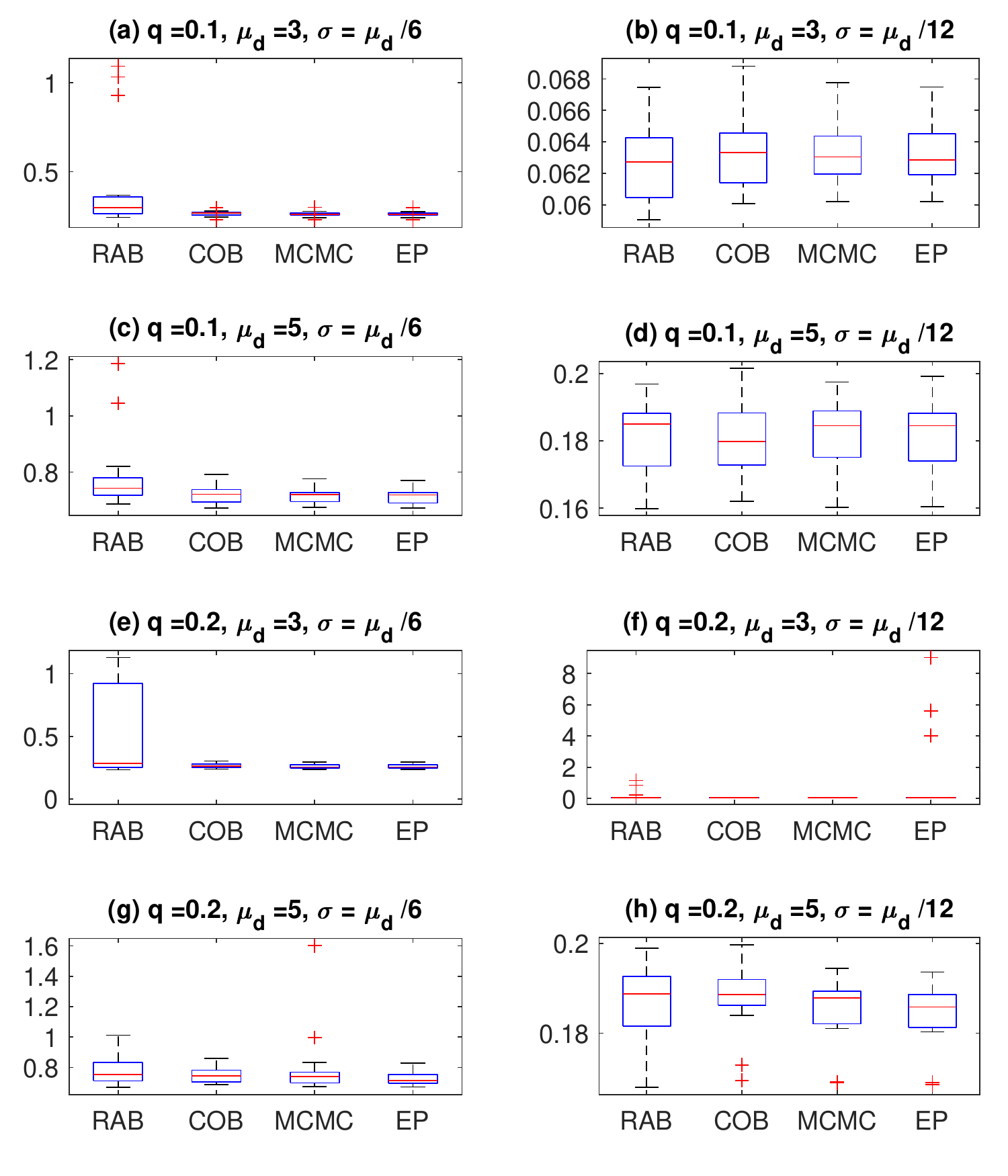}
	\caption{MSE performance of the four methods compared for eight simulated scenarios with 1D synthetic dataset. $q$ is the fraction of outliers in the training data, $\mu_d$ is the mean bias of outliers, and $\sigma$ is the standard deviation of data for both normal data and outliers.}
	\label{Figure4}
\end{figure}

We did more quantitative analysis on the 15 replicated outcomes of MSE, NLPD and TIME for each of the eight simulation scenarios. The MSE performance is summarized using the bar charts in Figure \ref{Figure4}. The average MSE values over 15 replicated outcomes are very close among the four methods. The variations of the MSE values are significantly different. The random bias model showed larger variations over the other methods, mainly due to the existence of a few bad performing cases. Both of the \texttt{MCMC} and \texttt{EP} approaches were quite stable. However, the \texttt{MCMC} approach has shown a couple of significantly bad performing cases as shown in Figure \ref{Figure4}-(g), mainly due to MCMC sampling variations. The \texttt{EP} approach has shown three significantly bad performing cases as shown in Figure \ref{Figure4}-(f), which occurred when the \texttt{EP} converged slowly relative to the other good performance cases. The outcomes from the constant bias model were less varied and more consistent than those from the other methods.

\begin{figure}[p]
	\includegraphics[width=\textwidth]{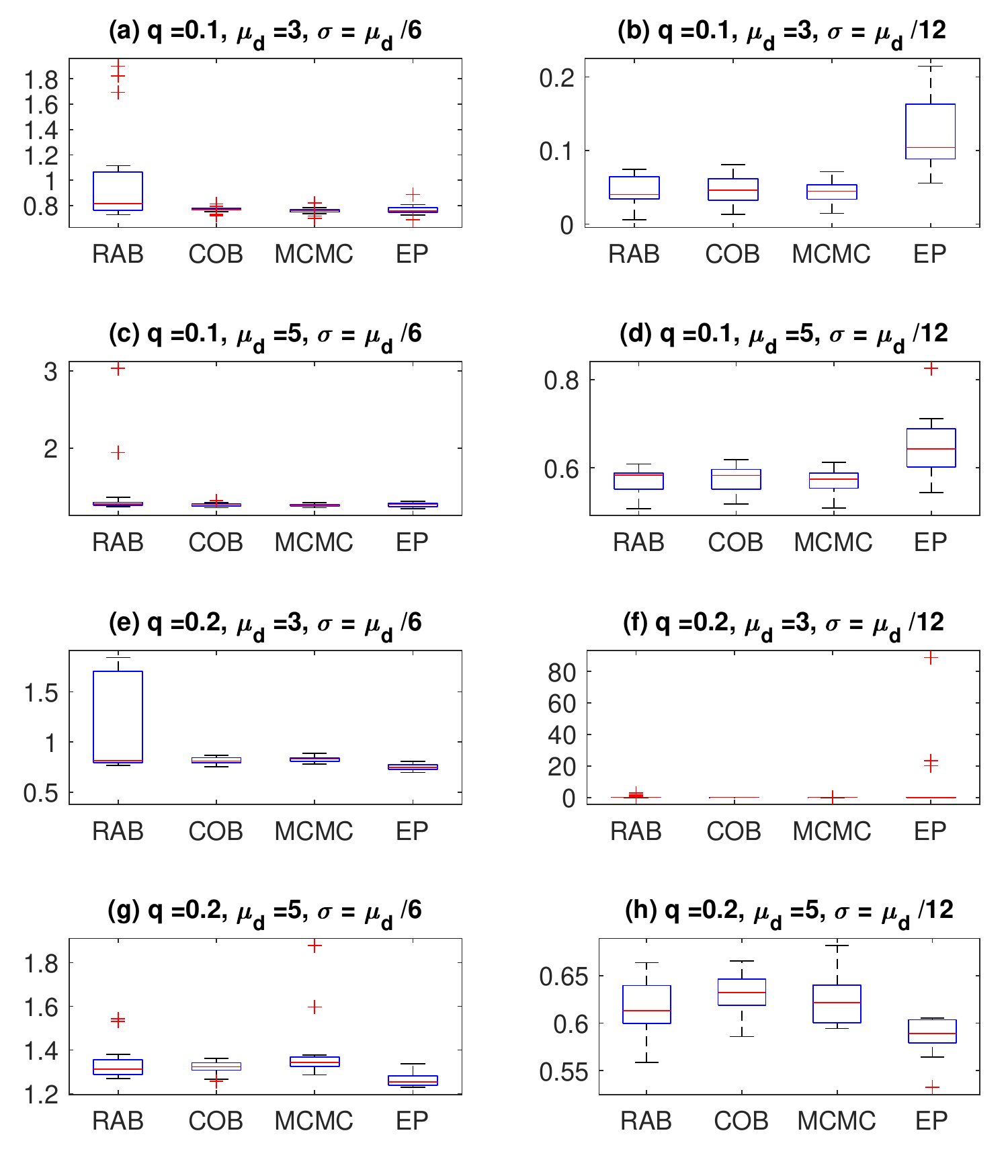}
	\caption{NLPD performance of the four methods compared for eight simulated scenarios with 1D synthetic dataset. $q$ is the fraction of outliers in the training data, $\mu_d$ is the mean bias of outliers, and $\sigma$ is the standard deviation of data for both normal data and outliers.}
	\label{Figure5}
\end{figure}

The NLPD performance was similar to the MSE performance except for that the \texttt{EP}'s average NLPD values are significantly higher than the corresponding values for the other compared methods. For example, see Figure \ref{Figure5}-(b) and \ref{Figure5}-(d). In the two scenarios, the MSE performance of the \texttt{EP} was similar to those of the other methods, so the differences in the NLPD are mainly due to the differences in the predictive variance estimates. As we illustrated in Figure \ref{Figure7}, the \texttt{EP} underestimates the predictive variances for some scenarios. Besides, the variations in the NLPD value for the \texttt{EP} are larger for these two scenarios. Again, the random bias model has shown relatively larger variations in the NLPD values for a couple of the simulated scenarios. 

The computation times of the four methods are quite different. The \texttt{MCMC} approach takes more computation times than the other methods, and the \texttt{EP} is the second slowest. Our proposed constant bias model is fastest among the four methods. The constant bias model provides a very efficient robust GP regression solution with great and less varying MSE and NLPD performances. This is the major benefit of using the proposed constant bias model. 

\begin{figure}[p]
	\includegraphics[width=\textwidth]{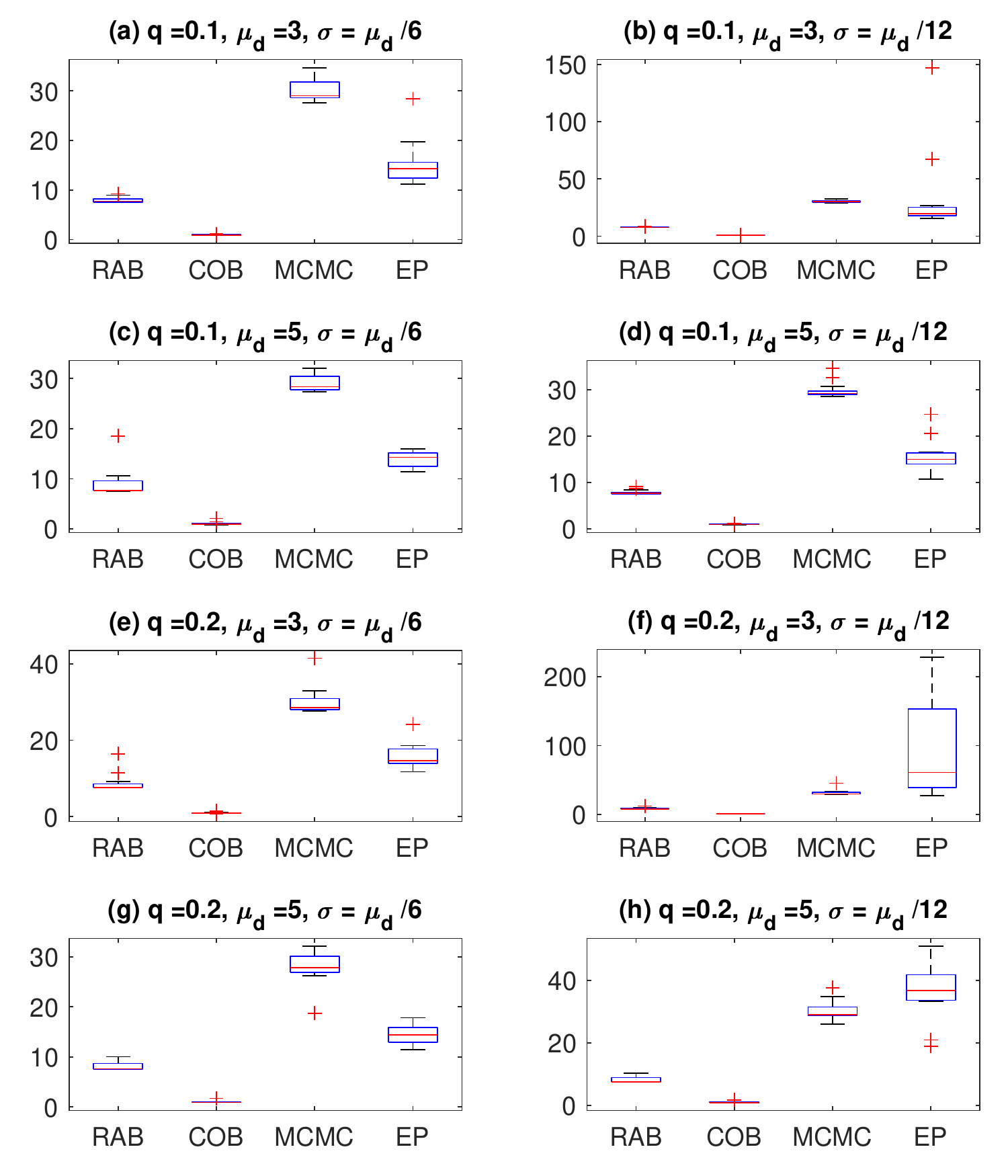}
	\caption{Time performance of the four methods compared for eight simulated scenarios with 1D synthetic dataset. $q$ is the fraction of outliers in the training data, $\mu_d$ is the mean bias of outliers, and $\sigma$ is the standard deviation of data for both normal data and outliers.}
	\label{Figure6}
\end{figure}

\subsection{10D Friedman Dataset}
We use this example to see how the four compared methods perform for high dimensional inputs. The Friedman data has 10 input dimensions, five of which are not related to the response variable but adds complexity in data analysis. The main interest here is how the accuracy and computation time of the four compared methods are affected by the increased input dimension. Like in the previous example, we collected the MSE, NLPD and TIME metrics for the four compared methods. 

\begin{figure}[p]
	\includegraphics[width=\textwidth]{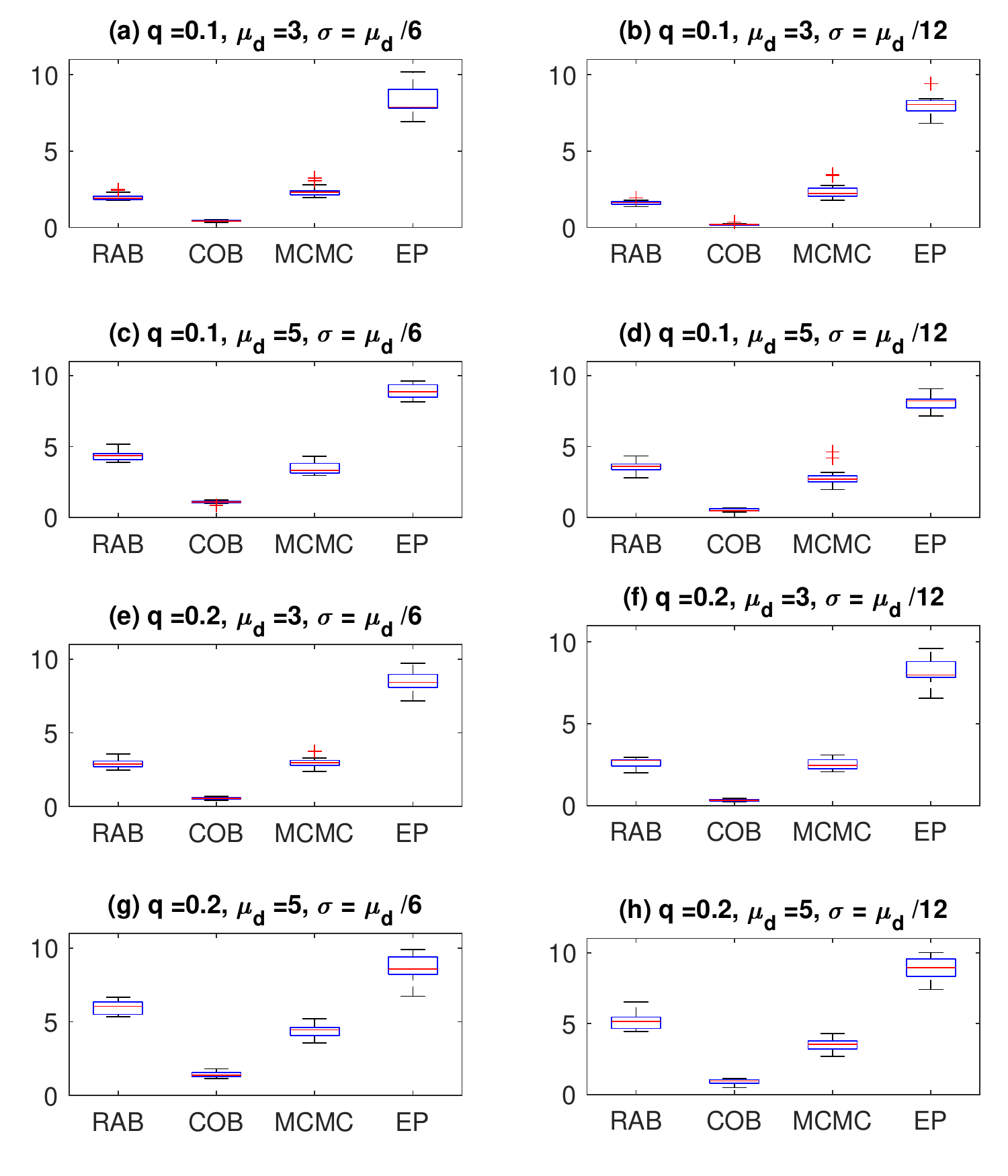}
	\caption{MSE performance of the four methods compared for eight simulated scenarios with 10D Friedman dataset. $q$ is the fraction of outliers in the training data, $\mu_d$ is the mean bias of outliers, and $\sigma$ is the standard deviation of data for both normal data and outliers.}
	\label{Figure8}
\end{figure}

Figure \ref{Figure8} summarizes the MSE values of the four methods for eight test scenarios. Different from the 1D dataset, the averages and variations of the MSE values are significantly different among the four compared methods. Our proposed constant bias model (\texttt{COB}) was the best performer for all of the eight scenarios with significant gaps to the other methods, and its average MSE values are the lowest while the variations of the MSE values are relatively little.  Our proposed random bias model (\texttt{RAB}) was the second-best consistently, and the \texttt{MCMC} approach was the third-best. The \texttt{EP} approach has shown a significant deterioration of the performance as the input dimension increases. A major benefit of using the \texttt{COB} is that the MSE performance changes much less in between the 1D dataset and this dataset. For example, its average MSE performance with the 1D data was 0.2649 while that with this 10D dataset was 0.4208 for $q=0.1$, $\mu_d = 3$ and $\sigma = \mu_d / 6$, but the gaps are much more significant for the other three methods as shown in Table \ref{Table1}.

\begin{table}
	\centering
	\begin{tabular}{|c|c|c|c|c|}
		\hline
		 				  & \texttt{RAB} & \texttt{COB} & \texttt{MCMC} & \texttt{EP} \\ \hline
		1D Synthetic Data & 0.2994       &  0.2674      & 0.2601        & 0.2598      \\
		10D Friedman Dataset & 1.9074   & 0.4209  &  2.2994  &  7.8828 \\
		\hline
	\end{tabular}
	\caption{Average MSE values for $q=0.1$, $\mu_d = 3$ and $\sigma = \mu_d / 6$}
	\label{Table1}
\end{table}

\begin{figure}[p]
	\includegraphics[width=\textwidth]{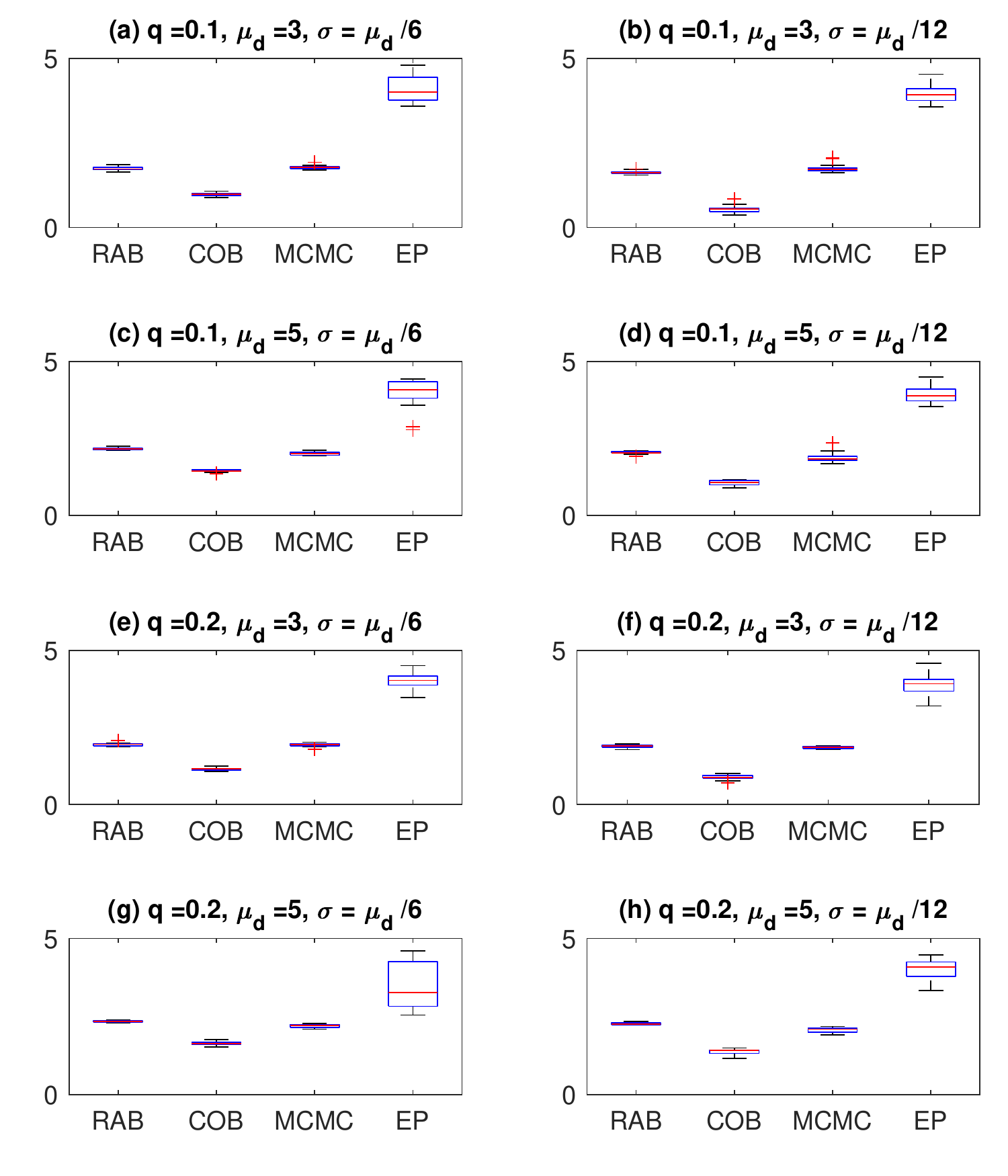}
	\caption{NLPD performance of the four methods compared for eight simulated scenarios with 10D Friedman dataset. $q$ is the fraction of outliers in the training data, $\mu_d$ is the mean bias of outliers, and $\sigma$ is the standard deviation of data for both normal data and outliers.}
	\label{Figure9}
\end{figure}

Figure \ref{Figure9} compares the NLPD performances of the four methods. The NLPD performance is quite comparable to the MSE performance. The proposed \texttt{COB} is the best performer with the lowest average negative log posterior density values and little variations. The NLPDs of the \texttt{RAB} and \texttt{MCMC} were comparable.

\begin{figure}[p]
	\includegraphics[width=\textwidth]{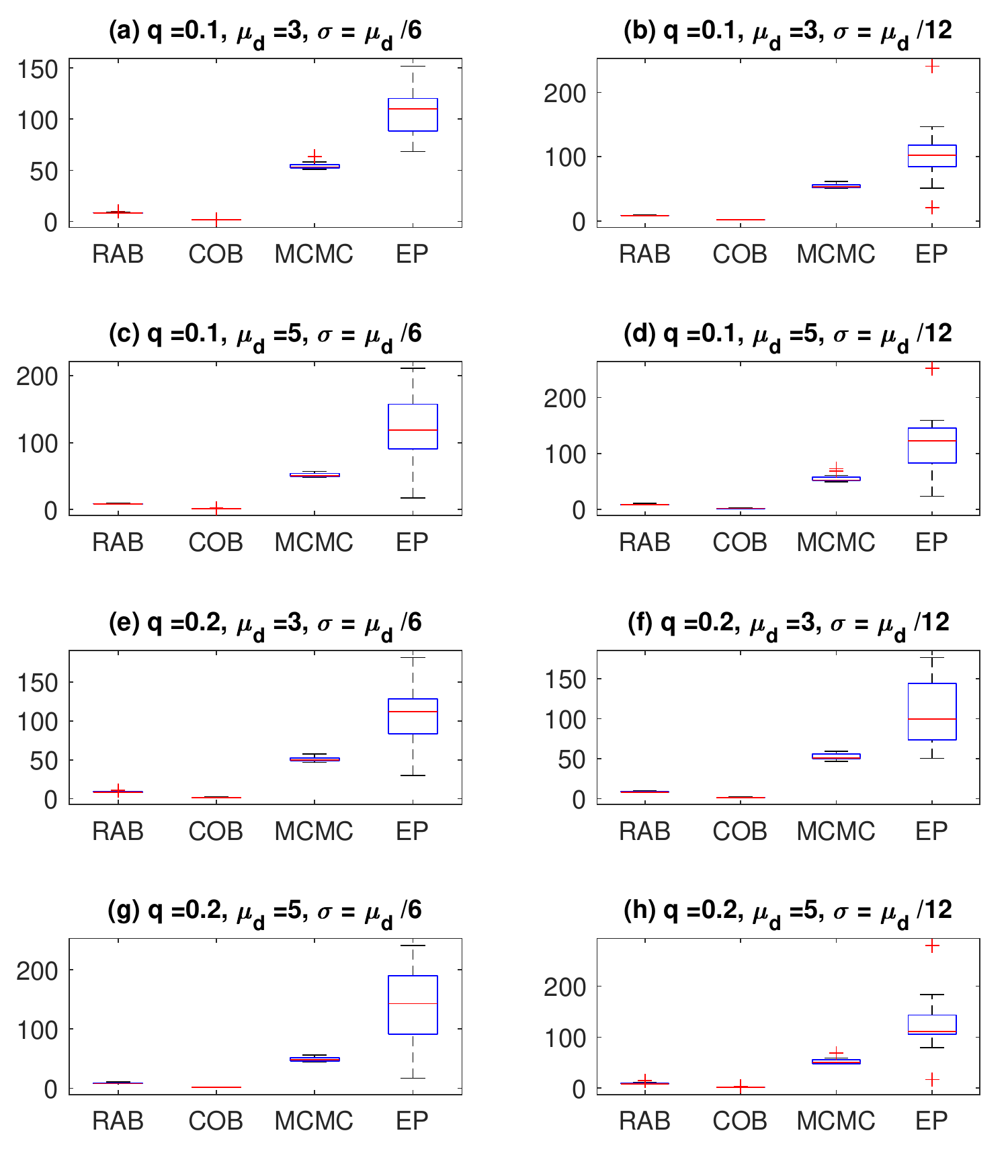}
	\caption{Time performance of the four methods compared for eight simulated scenarios with 10D Friedman dataset. $q$ is the fraction of outliers in the training data, $\mu_d$ is the mean bias of outliers, and $\sigma$ is the standard deviation of data for both normal data and outliers.}
	\label{Figure10}
\end{figure}

Figure \ref{Figure10} compares the total computation times. The \texttt{COB} is fastest. It is interesting to see that the \texttt{EP} slowed down for this high dimensional example significantly, being slower than the \texttt{MCMC}. Besides, the variation in the computation times was very significant for the \texttt{EP}, mainly because the convergence of the expectation propagation algorithm varied significantly over 15 random replicated experiments. 

The main message from the two simulated studies is clear. The predictive power of the proposed \texttt{COB} method is very competitive and stable, which does not depend significantly on training samples and input dimensions of the dataset. Its computation time is faster than the existing approximation methods by more than one order of magnitude, and it is also less dependent on the input dimension.

\section{Real Examples – Measurement of Environmental Variables} \label{section4}
\begin{figure}[ht!]
	\centering
	\includegraphics[width=\textwidth]{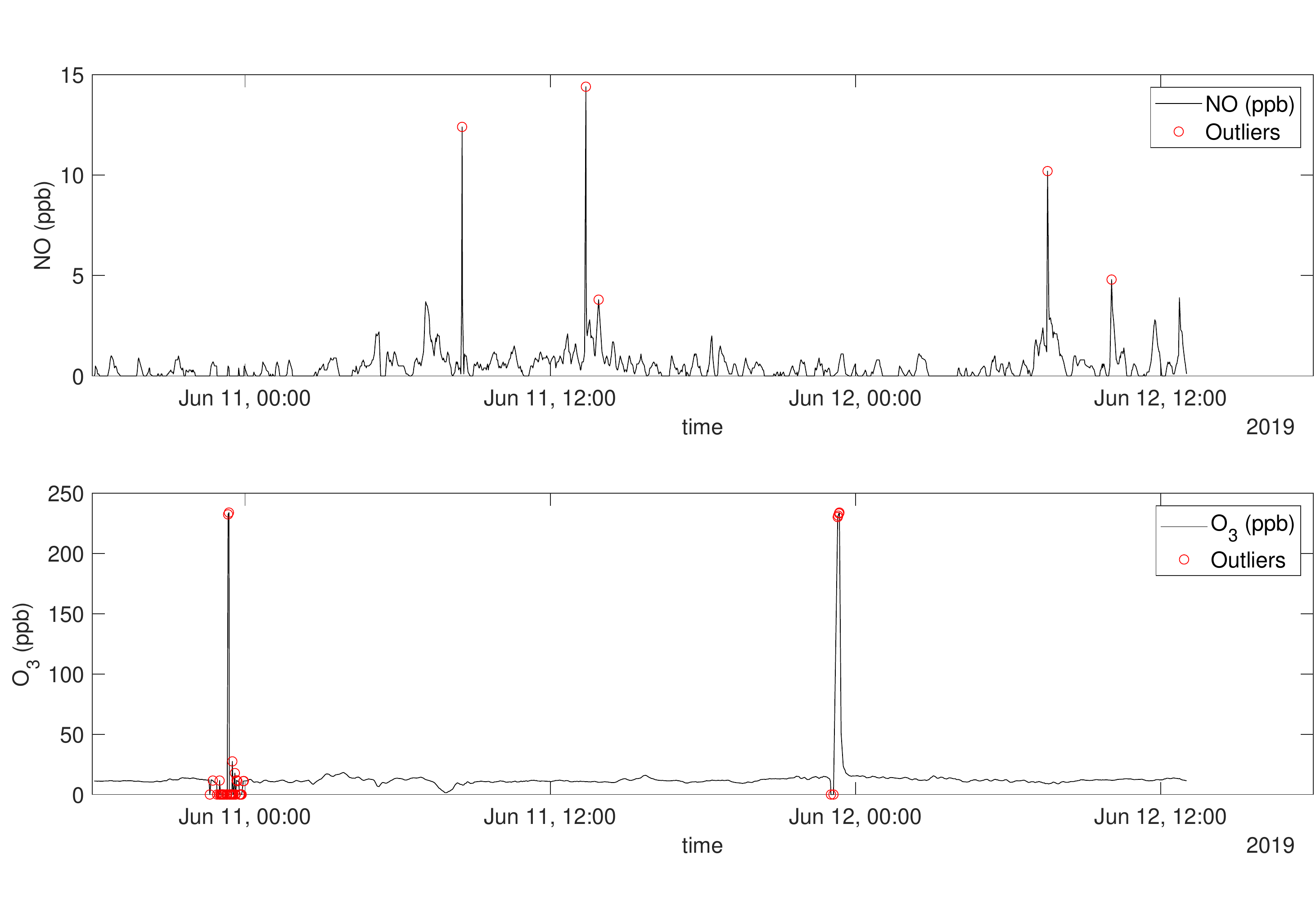}
	\caption{Environmental measurement data}
	\label{Figure11}
\end{figure}

In this section, we use environmental data collected by two monitoring systems deployed by the U.S. Air Force for development of atmospheric corrosion models of aluminum alloys (AA). We considered temperature and humidity from the {CorRES}\textsuperscript{\texttrademark} sensing system, and nitrogen oxides (NO$_x$) and ozone (O$_3$) gas concentrations in parts per billion (ppb) from the {Airpointer}\textsuperscript{\textregistered} system. Both systems were placed at an outdoor environmental test site operated by the U.S. Naval Research Laboratory in Key West FL. In this study, the variables of time, ambient temperature, in degree Celsius, and relative humidity are used as three-dimensional predictors, and each of the gas concentrations are used as response variables. Multiple causes for the gas concentration outlier data were identified, including loss of communications from sensors, and data produced during prescribed user calibrations. The NO$_x$ data have more intermittent outliers, while the O$_3$ data contain more clustered outliers \ref{Figure11}. For the O$_3$, the outliers deviate strongly from the normal data. Although outliers in the gas concentration data are only a small fraction of the total data, a computationally efficient process for accounting for these outliers increases the utility of the environmental monitoring systems and data sets. The capability of the robust GP regression to account for response variable outliers in the gas monitoring data sets was determined relative to three other existing approaches for these two unique NO$_x$ and O$_3$ scenarios. 
 
Another major contrast from the simulated datasets is that this real dataset contains more measurements. In total, the dataset has 1,400 measurements for each of the gas concentrations. We manually identified outliers as indicated with red circles in Figure \ref{Figure11}. There are five outliers for the NO gas and 45 outliers for the O$_3$ gas. For each of the gas types, we randomly selected 10\% or 20\% of the 1,400 records excluding the outliers, which serve as a test dataset. So, the test dataset only contains normal data.  The training data contains the whole 1,400 records, substituting the records selected for the test dataset with outliers. The outliers are randomly sampled from those which we manually identified. Dropping our proposed random bias model (\texttt{RAB}) in this comparison due to its poor performance for this example, we evaluate the three other methods in terms of the three performance metrics, MSE, NLPD and TIME. In this study, we applied the exponential kernel for all of the four methods. 

\begin{figure}[ht!]
	\centering
	\includegraphics[width=\textwidth]{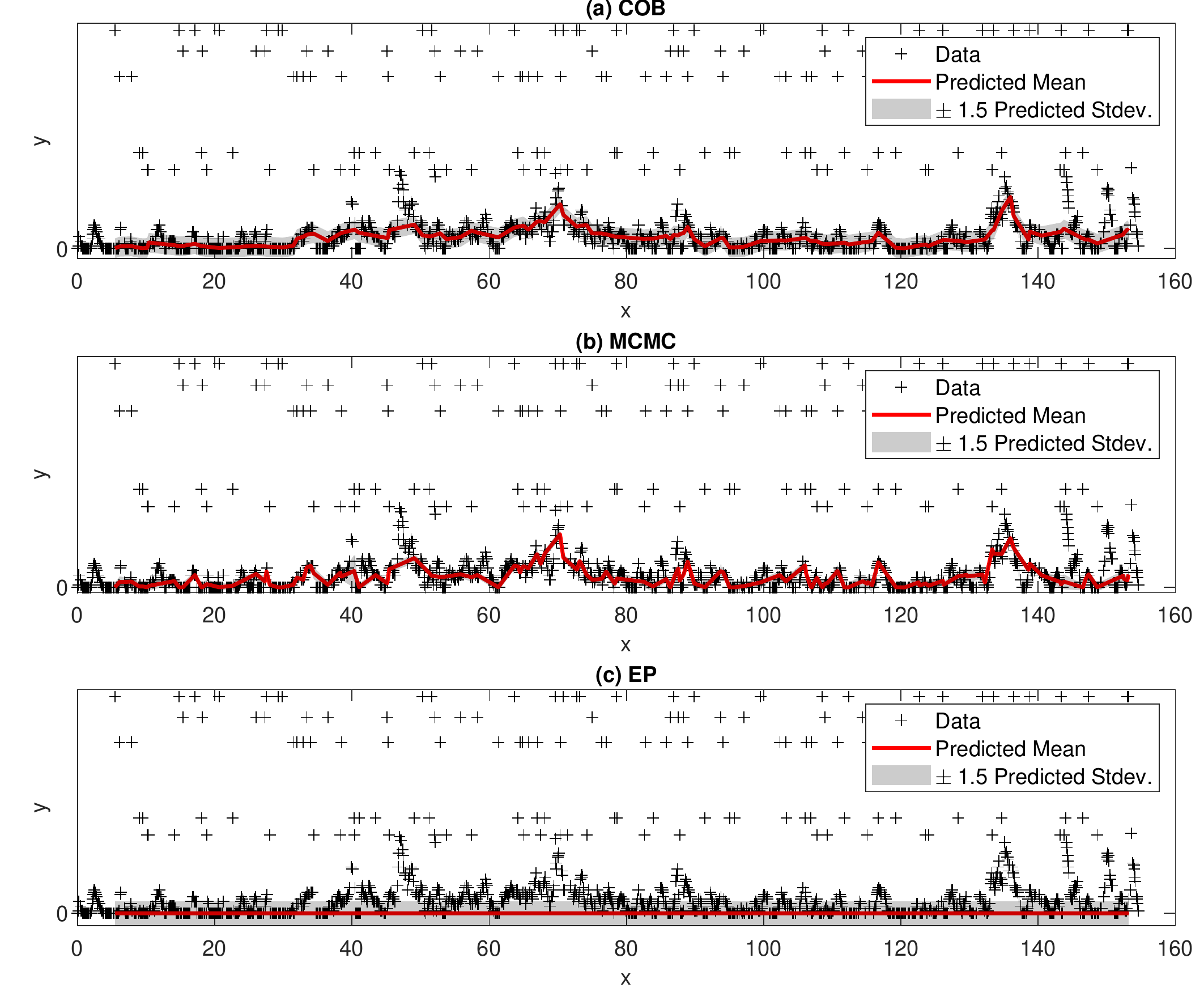}
	\caption{Predictive mean and variances of the three compared methods for environmental measurement data, NO measurements with 10\% outliers.}
	\label{Figure12}
\end{figure}

Table \ref{Table2} summarizes the outcomes with the MSE, NLPD and TIME evaluations. The \texttt{MCMC} approach worked best for the NO data, and the proposed \texttt{COB} model worked closely to the \texttt{MCMC} approach. However, there is a meaningful gap in between them. By looking at their predictive means and variances as shown in Figure \ref{Figure12}, we can see that both of the models work reasonably, but the \texttt{COB} model over-smooths the wiggly fluctuations of the NO measurements. The \texttt{EP} approach did not work very well, creating a flat predictive mean and too wide predictive variance around the mean. It is mainly due to the poor convergence of its hyperparameter optimization for this example. 

\begin{figure}[ht!]
	\centering
	\includegraphics[width=\textwidth]{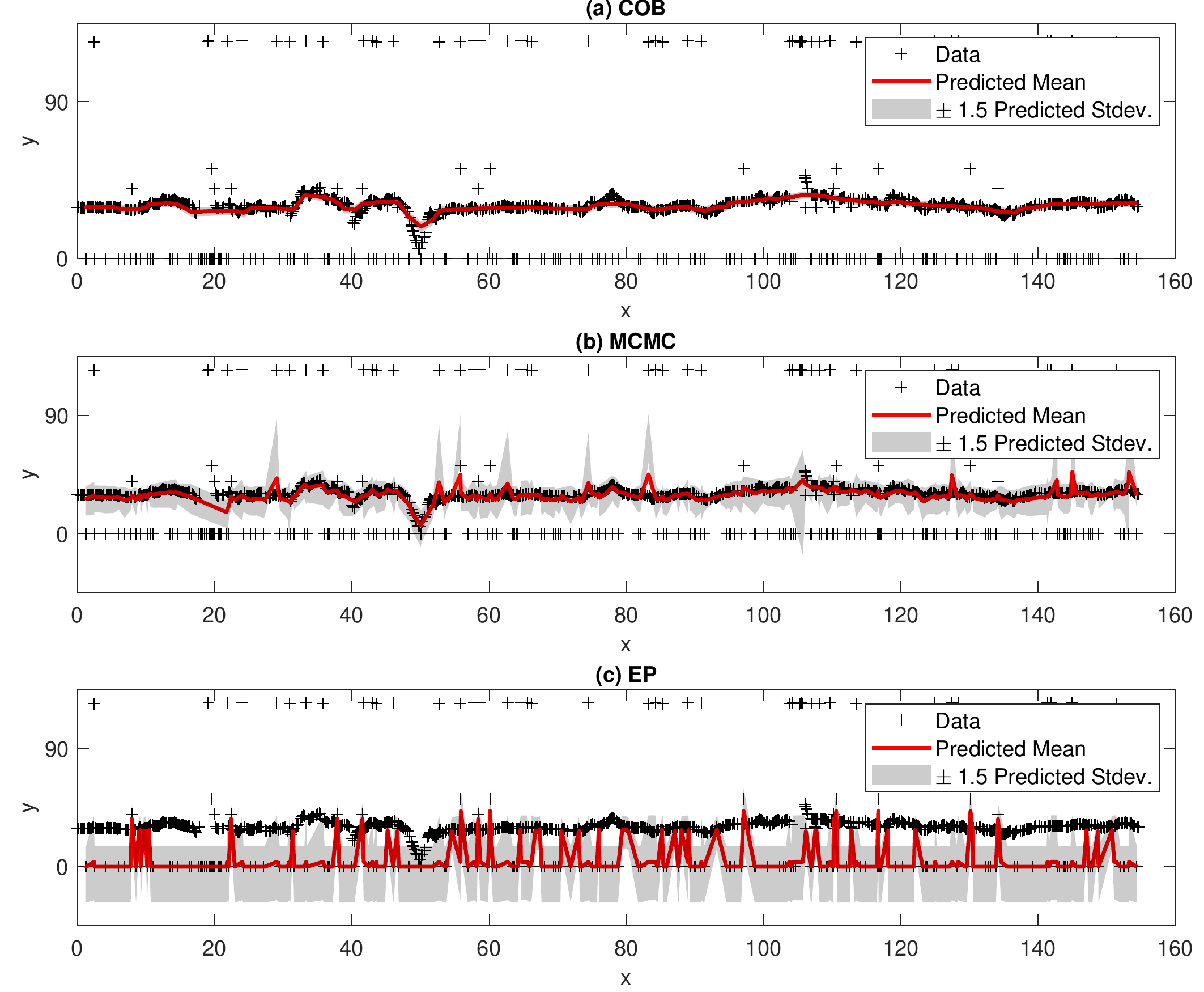}
	\caption{Predictive mean and variances of the three compared methods for environmental measurement data, O$_3$ measurements with 20\% outliers.}
	\label{Figure13}
\end{figure}

For the O$_3$ measurement, the proposed \texttt{COB} model worked better in MSE, but the \texttt{MCMC} approach worked better in NLPD. Figure \ref{Figure13} shows the predictive means and variances of the three compared methods.  Our proposed model better catches the overall trend of normal data, and the \texttt{MCMC} approach is influenced by outliers for some test locations. The \texttt{EP} failed to find the mean line of the normal data, catching up with the outlier data below the normal mean line. 

All in all, the proposed \texttt{COB} model works better than or close to the best performing method for all of the simulated and real data cases with much shorter computation time. The proposed model can be a very time economic option for the robust GP regression with great prediction accuracy and robustness.  

\begin{table}
	\centering
	\begin{tabular}{|c|c|c|c|c|}
		\hline
		Dataset (Percentage Outliers) & Metrics &  \texttt{COB} & \texttt{MCMC} & \texttt{EP} \\ \hline
		NO (10\%) 				      & MSE     &  0.1104 &  0.0436  &  0.5024  \\
		                              & NLPD    &  0.3605 &  -0.7021 &  2.0554  \\ 
		                              & TIME    &  51 Sec &  1,070 Sec & 206 Sec \\ \hline
		NO (20\%) 				      & MSE     &  0.1974 &   0.0978 &   0.5267  \\
									  & NLPD    &  0.6078 &  0.2193  &  1.3589  \\ 
									  & TIME    &  53 Sec &  961 Sec &  175 Sec \\ \hline
		O$_3$ (10\%)			      & MSE     &  9.2467 &   9.3090 & 160.2542  \\
									  & NLPD    &  13.1931 &    6.4755  &  3.9824  \\ 
									  & TIME    &  40 Sec &  983 Sec &  168 Sec \\ \hline
		O$_3$ (20\%)			      & MSE     &  5.5220 &   8.0989 & 126.9505  \\
									  & NLPD    &  4.3783 &  2.2006  &  6.8298  \\ 
									  & TIME    &  50 Sec &  833 Sec &  991 Sec \\ 
		\hline
	\end{tabular}
	\caption{MSE, NLPD, TIME values for environmental measurement data}
	\label{Table2}
\end{table}

\section{Conclusion} \label{section5}
We presented a new computationally efficient method to a robust GP regression. The approach regards data as noisy and biased observations of an underlying regression function. Accordingly, the likelihood linking data to the regression function includes bias terms. We modeled the bias terms in two different ways, biases as unknown fixed parameters and biases as unknown random quantities. We estimated the parameters of the resulting bias models based on the proposed regularized maximum likelihood estimation. We entailed how the regularized likelihood functions are formulated and how the tuning parameters are chosen. After the bias model estimation, the robust GP regression can be achieved following the standard GP procedure. This bias modeling approach is very simple and computationally efficient, while it provides excellent prediction accuracy and robustness to outliers. We examined its numerical performance based on a comprehensive simulation study. For most of the simulation cases, the proposed approach outperformed the existing approaches of the fractional EP approximation with the Student-t likelihood and the MCMC approach with the Gaussian scale mixture likelihood. The performance of the proposed approach was also less sensitive to the proportion of outliers and the noise level of data as well as the dimension of data. The computation time of the proposed approach was faster than the existing approaches by at least one order of magnitude. The new approach has also shown better trade-offs between the computational time and prediction accuracy than the existing approaches in our numerical study with environmental sensor data. We believe that the proposed bias model is a very attractive solution to a robust GP regression. 

%\section*{Appendix. Useful formula}
%\begin{equation*}
%\begin{split}
%& p(y|f) = \mathcal{N}(Af, K) \\
%& p(f) = \mathcal{N}(\mu, B) \\
%& p(y) = \mathcal{N}(A\mu, K+ABA^T) \\
%& p(f|y) = \mathcal{N}((A^TK^{-1}A+B^{-1})^{-1} (A^TK^{-1}y + B^{-1}\mu)  , (A^TK^{-1}A+B^{-1})^{-1})
%\end{split}
%\end{equation*}

% Acknowledgements should go at the end, before appendices and references

\acks{We would like to acknowledge support for this work from Air Force Research Laboratory Materials and Manufacturing Directorate and members members of the AFRL Corrosion Integrated Product Team. The work was conducted under U.S. Federal Government Contract No FA8650-15-D-5405.}


%\newpage
%\appendix
%\section*{Appendix A.}

\bibliography{gp}

\end{document}